\documentclass[sigplan,screen]{acmart}
\AtBeginDocument{%
  }

\setcopyright{acmlicensed}
\copyrightyear{2025}
\acmYear{2025}

\usepackage{rotating}

\begin{document}

\title{Beyond Test-Time Compute Strategies: Advocating Energy-per-Token in LLM Inference}

\author{Patrick Wilhelm}
\email{patrick.wilhelm@tu-berlin.de}
\affiliation{%
  \institution{BIFOLD}
  \institution{Technische Universität Berlin}
  \city{Berlin}
  \country{Germany}
}
\author{Thorsten Wittkopp}
\email{t.wittkopp@tu-berlin.de}
\affiliation{%
  \institution{Technische Universität Berlin}
  \city{Berlin}
  \country{Germany}
}

\author{Odej Kao}
\email{odej.kao@tu-berlin.de}
\affiliation{%
  \institution{Technische Universität Berlin}
  \city{Berlin}
  \country{Germany}
}

\renewcommand{\shortauthors}{Wilhelm et al.}

\begin{abstract}
Large Language Models (LLMs) demonstrate exceptional performance across diverse tasks but come with substantial energy and computational costs, particularly in request-heavy scenarios. In many real-world applications, the full scale and capabilities of LLMs are often unnecessary, as Small Language Models (SLMs) can provide accurate responses for simpler text generation tasks. When enhanced with advanced reasoning strategies, such as Chain-of-Thought (CoT) prompting or Majority Voting, SLMs can approach the performance of larger models while reducing overall computational requirements. However, these strategies can also introduce additional energy costs, creating an energy-accuracy trade-off. Our analysis examines these trade-offs in test-time compute strategies for smaller models compared to larger ones, using the MMLU benchmark. Additionally, we explore the input-output token dynamics of transformer architectures, which result in nonlinear hardware energy operation curves for LLMs. To bridge AI research with its physical impact, we propose \textit{energy efficiency metrics}, including Energy-per-Token, as complements to traditional accuracy benchmarks. Beyond model selection, we propose controlled reasoning in CoT token generation, using operating curves to regulate reasoning depth dynamically. This vision integrates a energy-aware routing mechanism, ensuring that model selection and inference strategies balance accuracy for sustainable AI deployment.
\end{abstract}



\received{10 February 2025}

\maketitle

\section{Introduction}

As AI models become more accessible and integrated into IT systems, sustainability and computational concerns are growing. 
Recent research shows, that the energy demand of global data centers is projected to reach 1,000 TWh by 2026~\footnote{IEA Electricity Report 2024, https://www.iea.org/reports/electricity-2024}, fueled in part by the rapid expansion of AI technologies. 
Additionally, related carbon emissions are estimated to account for up to 8 \% of global emissions in the next decade~\cite{designingSustainableCompSystems}. 
The energy demand of Large Language Models (LLMs) are primarily driven by their parametric size and computational requirements.
Neural scaling laws, initially introduced by Kaplan~\cite{kaplan2020scaling}, offered a foundational framework for optimizing model performance by balancing model parameter size, dataset scale, and computational power, leading to a constant model parameter size increasement.

However, these scaling laws often neglect inference, a phase with different and task-specific computational demands. 
Research from cloud providers such as AWS and Google confirms that inference frequently surpasses training in energy consumption, especially in high-demand, low-latency applications~\cite{samsi2023words}. 
Although these frameworks have guided efficient training practices, only recent work accounts for the inference demand in the training phase, moving towards development of Small Language Models~\cite{sardana2023beyond}.
Recent research breakthrough in cost efficient LLM training by DeepSeek does not necessarily reduce resource consumption - instead, it often drives greater adoption and usage, known as the Jevons paradox.\cite{guo2025deepseek}.

This emphasizes the need to evaluate computing efficiency during inference alongside traditional metrics like accuracy \cite{luccioni2023counting,luccioni2023estimating,luccioni2025efficiencygainsreboundeffects,luccioni2024power}.
One might expect that Energy Consumption per Token during inference would be consistent for models with identical parameter sizes. 
In this paper, we show that they are, in fact, \textit{not}: LLMs reveal different \textit{energy efficiencies} across the same tasks on the same hardware. 
Our key findings include the following:
\\

\begin{itemize}
    \item Different LLM architectures have different energy efficiencies in token processing.
    \item LLM computations are influenced by the transformer-based autoregressive generation of tokens. These dynamics produce characteristic nonlinear energy operating curves over the number of generated tokens across LLMs.
    \item Test-time strategies like Chain-of-Thought Prompting boost accuracy in smaller models but come with high energy costs, making larger models more energy efficient in comparison.
\end{itemize}

Our findings are based on several experiments, detailed throughout the remainder of this paper.
After discussing the relevant related work in Section \ref{sec:background}, we provide a brief overview of the evaluated LLMs, the datasets, and the benchmarks used in Section \ref{sec:datasetsmodelshardware}. 
In Section \ref{sec:experiments}, we investigate energy efficiency of LLM during inference, highlighting how prompts lead to differences in energy consumption between models. 
We further characterize these variations by analyzing the input-output token relationships and their impact on the hardware energy operating curves. Lastly, in Section \ref{sec:discussion}, we propose a routing design balancing accuracy and sustainability in LLM inference.

\section{Balancing Compute in LLM Training and Inference}
\label{sec:background}
Scaling laws provide a structured framework for optimizing the balance between model size, training data, and computational resources. Kaplan et al.\cite{kaplan2020scaling} demonstrated that increasing model size significantly improves performance, shaping the development of large-scale language models. Building on this, Hoffmann et al.\cite{hoffmann2022training} refined these principles with the Chinchilla scaling laws, advocating for a proportional increase in model parameters and dataset size to achieve optimal training efficiency.
While Chinchilla emphasized efficiency in pretraining, recent models such as LLaMA 2 and LLaMA 3 have adopted a different strategy, prioritizing extensive training token counts—2 trillion and 15 trillion tokens, respectively—over major architectural changes~\cite{touvron2023llamaA,touvron2023llamaB}. This shift highlights a trade-off where higher training costs are offset by reduced inference costs and greater adaptability to deployment scenarios~\cite{sardana2023beyond}. Additionally, emerging research suggests that further refinements to scaling laws may integrate considerations of both model quality and inference demands, optimizing LLM designs for specific operational requirements~\cite{sardana2023beyond}.

Optimizing LLMs requires balancing computational demands between training and inference. While early scaling approaches focused on maximizing model performance during training, recent research extends these laws to consider inference efficiency~\cite{sardana2023beyond}.
Compression techniques, including quantization and pruning, are commonly employed to mitigate inference costs. Quantization reduces model precision while maintaining accuracy~\cite{jacob2018quantization, lin2024awq}, and pruning eliminates redundant parameters, lowering computational complexity~\cite{han2015deep, sun2023simple}. These techniques enable smaller, more efficient models that preserve performance while reducing resource demands. Moreover, advancements in test-time computation, such as dynamic inference optimization, continue to shape the trade-offs between training and inference efficiency, pointing to a future where these aspects become increasingly intertwined~\cite{epoch2023tradingoffcomputeintrainingandinference,snell2024scaling}.

Advances in inference-time optimization improve efficiency without requiring larger models. Chain-of-Thought (CoT) prompting enables stepwise reasoning, improving performance on complex tasks~\cite{wei2022chain, wang2024chain}. Variants such as Majority Voting and Best-of-N further enhance accuracy by generating multiple responses and selecting the most probable one~\cite{lightman2023let, uesato2022solving, wang2022self}.
More advanced inference-time strategies dynamically allocate computational resources. Beam Search expands multiple reasoning paths in parallel, selecting the most probable sequence based on cumulative likelihood. Monte Carlo Tree Search (MCTS) further refines this by exploring solution paths, evaluating their expected quality, and backpropagating optimal results~\cite{feng2023alphazero, wang2023math, xin2024deepseek}. Despite these improvements, most studies do not report computational costs or energy usage during inference, leaving a critical gap in understanding inference efficiency. The absence of standardized metrics for inference-time compute further complicates direct comparisons between different optimization techniques.
Transformers, the backbone of LLMs, pose additional computational challenges. The self-attention mechanism scales quadratically with sequence length, increasing memory and energy consumption~\cite{vaswani2017attention, rajbhandari2022deepspeed}. Sequential decoding exacerbates this problem, as token generation requires recurrent attention to past outputs, leading to significant computational overhead.

Efforts to improve LLM efficiency extend beyond model-level optimizations to system-level strategies. Techniques such as vLLM and Orca reduce inference memory footprints through continuous batching and paging, improving energy efficiency~\cite{kwon2023efficient, stojkovic2024dynamollm}.
Model partitioning strategies, such as those implemented in Clover, enhance deployment efficiency by distributing computational loads~\cite{li2023clover}. Sustainability-focused initiatives like Sprout minimize carbon emissions by reducing generated token counts without compromising answer quality~\cite{li2024sustainablegenaiusinggeneration}. By focusing on optimizing both encoding and decoding processes, these methods contribute to reducing overall energy consumption in large-scale LLM inference. However, standardized reporting on energy efficiency remains limited, making it difficult to assess the full impact of these optimizations~\cite{wilkins2024offline}.

\section{Datasets, Models and Hardware}
\label{sec:datasetsmodelshardware}
We focus on studying energy consumption in autoregressive token generation of different LLM architectures, specifically analyzing energy per token metrics, equation~\ref{eq:energyefficiency}. 
For our experiments, we used the datasets MMLU and MT-Bench.
In MMLU, all LLMs generate only a single output token per query, whereas for MT-Bench, we evaluate the metrics for token generations to account for varying computational demands.

\begin{itemize}
    \item \textbf{MT-Bench~\cite{zheng2023judging}}: A dataset for evaluating LLMs in multi-turn dialogues with 80 open-ended prompts across eight domains, including math, coding, and writing. 
    \item \textbf{Massive Multitask Language Understanding \\ (MMLU)\cite{hendrycks2020measuring}}: The MMLU dataset is a benchmark designed to evaluate the multitask accuracy of language models. It contains 57 categories spanning a wide range of tasks, including humanities, STEM, social sciences, and other professional domains. We clustered those categories into \cite{jaiswal2023compressing}: Computer Science, Math, Natural Sciences, Economics, Humanities, Health, Sociology and Engineering. 

\end{itemize}



All experiments are done on a NVIDIA L40S. For measuring the energy consumption, the python wrapper of the NVIDIA Management Library, which is also used in the codecarbon project \cite{benoit_courty_2024_11171501}. No parallelism is applied.
Batching impacts AI inference energy efficiency by defining how many data samples are processed simultaneously. Larger batches improve hardware utilization, but for consistency, we set the batch size to 1 for all models. This ensures fair, reliable performance comparisons under identical conditions.

\section{Evaluating LLM Energy Efficiency}
\label{sec:experiments}
In this section, we analyze the energy efficiency of various LLMs when processing identical inputs on the same hardware. Our focus is on measuring the energy consumption of Transformer based LLM architectures during both input processing and sequential token generation, which we control by setting a fixed maximum token output. Additionally, we assess the impact of applied reasoning techniques using a small language model (SLM), examining their effects on both accuracy and energy efficiency. \\We investigate following research questions:\\
\begin{itemize}
    \item \textbf{RQ1: How does energy efficiency in LLMs differ for the same input?}\\
    \item \textbf{RQ2: How does sequential output token generation impact energy consumption?}\\
    \item \textbf{RQ3: How do test-time compute strategies, such as Majority Voting and Chain-of-Though Prompting, affect the trade-off between accuracy and energy consumption?}
\end{itemize}

For Research Question 1, we assess the efficiency of various 7B-parameter LLMs in processing the MT-Bench dataset while generating a single token, achieved by setting the max new tokens parameter to 1. For Research Question 2, we conduct multiple runs to analyze the impact of sequential token generation on energy consumption. Finally, we apply Chain-of-Thought reasoning to the 1B LLaMA 3.2 model and compare its accuracy and energy consumption against the zero-shot performance of both the vanilla 1B model and the LLaMA 8B.

The equation \ref{eq:energyefficiency} Energy Efficiency quantifies the energy efficiency of a  model during the processing and generation of tokens. 
The formula is given by:

\begin{equation}
\text{Energy per Token [Joule]} = \frac{W_{\text{consumed}}* Time(s)}{T_{\text{processed}}}
\label{eq:energyefficiency}
\end{equation}

\begin{itemize}
    \item \( W_{\text{consumed}}* Time(s) \) represents the \textbf{total energy} consumed by the GPU during the entire processing and generation phase. This includes the power (in Watts) used by the hardware to process the input data and generate the corresponding output.
    \item \( T_{\text{processed}} \) is the \textbf{total number of tokens} processed, which is the sum of both the input tokens (the tokens fed into the model) and the output tokens (the tokens generated by the model as a response).
\end{itemize}

\subsection{Energy Efficiency for Input Token Processing}
The MT-Bench dataset is utilized for input prompt processing. With the python nvidia management library the energy consumption of the hardware is measuring and the latency to process the prompt as well as generate the first output token. This allows us to isolate the effect of sequential autoregression in transformer networks and focus solely on comparing different architectures for processing the same input prompts while generating a single output token.
\begin{figure}[h]
\centering
  \includegraphics[width=\columnwidth]{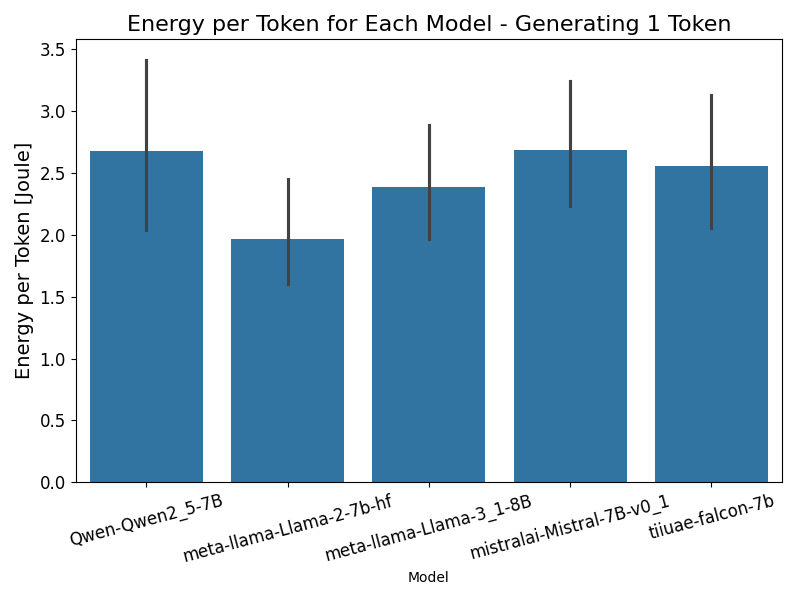}
  \caption{LLMs differ in token processing efficiency}
  \label{fig:efficiency_averagebarplot}
\end{figure}

\begin{figure}[h]
\centering
  \includegraphics[width=\columnwidth]{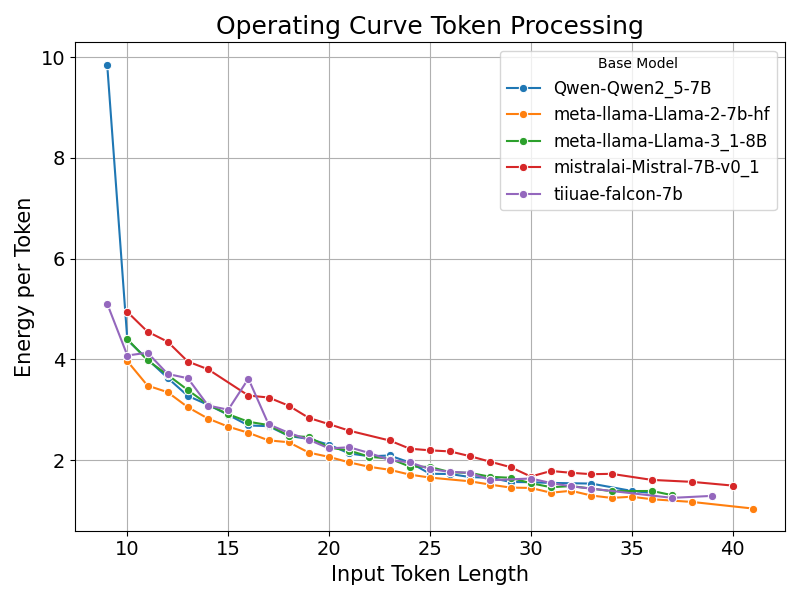}
  \caption{Non-linear behavior in token processing}
  \label{fig:inputtoken_operatingcurve}
\end{figure}

Figure \ref{fig:efficiency_averagebarplot} visualizes the average energy efficiency of different LLMs processing the MT-Bench. 
We can see that different architectures of Large Language Models consumed different amount of energy processing the same input. 
Extending this observation figure \ref{fig:inputtoken_operatingcurve} shows a unique characteristic for each LLM in processing different lengths of input token. 
A clear nonlinear behavior is observable, allowing to fit a function and to predict how much energy the processing of different amount of input token could consume.

This evaluation shows that for research question 1: \textit{How does energy efficiency in LLMs differ for the same input?} Its clearly identifyable that each model works with different efficiencies in input token processing.


\subsection{Energy Efficiency for Output Token Generation}
In transformer-based language models, token generation is inherently autoregressive, where each new token is generated sequentially.
As the model generates additional tokens, the computational complexity increases with each new output due to repeated application of attention layers. This increased complexity leads to variations in energy consumption in different architectures. 
In this study, we conducted an empirical evaluation of several LLM architectures under identical input conditions but varying numbers of generated tokens. By analyzing the energy efficiency of these models, we gain valuable insight into the operational efficiency and scalability of different transformer architectures during autoregressive token generation.
\begin{figure}[h]
\centering
  \includegraphics[width=\columnwidth]{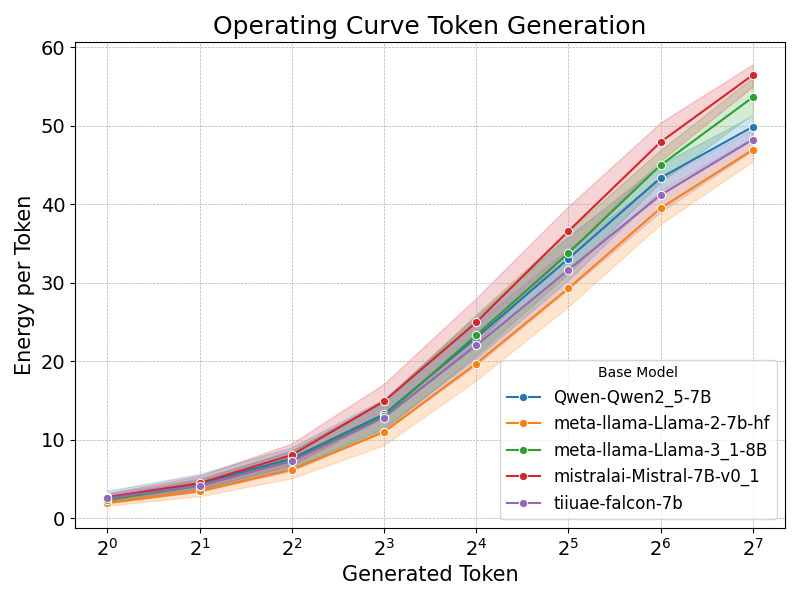}
  \caption{Non-linear behavior in token generation}
  \label{fig:operating_curves}
\end{figure}

\begin{figure}[h]
 \centering
   \includegraphics[width=\columnwidth]{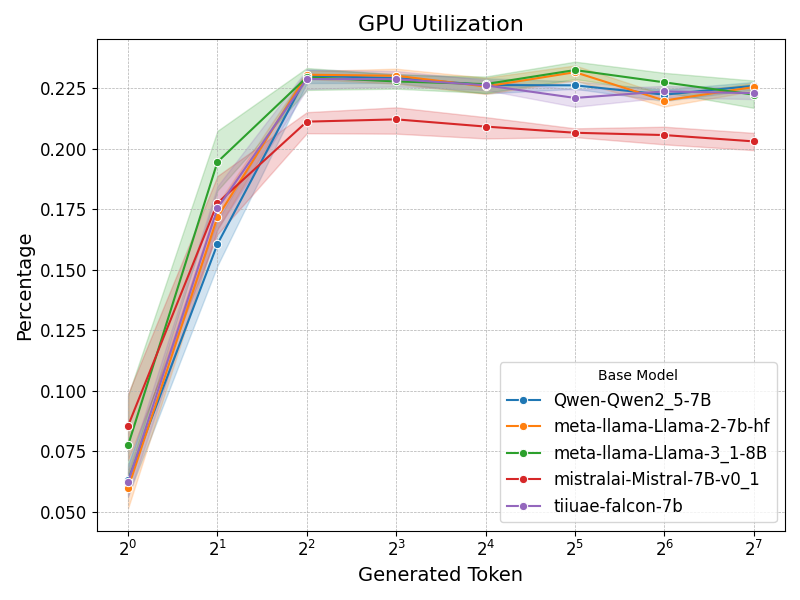}
   \caption{GPU utilization of different LLM architectures on the same Hardware.}
   \label{fig:operating_curves_gpuutil}
\end{figure}


Figure \ref{fig:operating_curves} illustrates energy consumption as a function of generated tokens, up to 128 tokens. Notably, all LLMs exhibit a nonlinear trend, differing in amplitude but following a similar general pattern. This suggests that while sequential token generation behavior is consistent across models, the intensity and magnitude vary.

This observation enables the fitting of parametric functions to approximate model-specific operational curves for energy consumption, which is crucial for the proposed routing architecture in Section \ref{sec:discussion}.

A potential indicator of this behavior is the GPU utilization. Figure \ref{fig:operating_curves_gpuutil} shows that while Mistral has the lowest GPU utilization, it also exhibits the highest energy consumption per token in Figure \ref{fig:operating_curves}. However, this pattern is not universal, as other model architectures share similar GPU utilization yet display distinct energy characteristics in their operating curves.

This evaluation shows that for research question 2: \textit{How does sequential output token generation impact energy consumption?} The sequential token generation has a characteristic impact on their energy per token metric. This nonlinear function is observable for all evaluated Transformer based Language Models, opening energy cost estimations given a model and specific hardware.

\subsection{Trade-Off between Accuracy and Energy Consumption}


\begin{table*}[htbp] 
\centering
\begin{tabular}{|c||c|c||c|c||c|c||c|c|}
\hline
{\textbf{Category}} & \multicolumn{4}{c||}{\textbf{Zero-Shot}} & \multicolumn{4}{c|}{\textbf{Reasoning}} \\
\cline{2-9}
& \multicolumn{2}{c||}{\textbf{Llama 1B}} & \multicolumn{2}{c||}{\textbf{Llama 8B}} & \multicolumn{2}{c||}{\textbf{Llama 1B MV}} & \multicolumn{2}{c|}{\textbf{Llama 1B CoT}} \\
\cline{2-9}
& \textbf{Acc.} & \textbf{Energy} & \textbf{Acc.} & \textbf{{$\Delta$\%}Acc, {$\Delta$\%}E} & \textbf{Acc.} & \textbf{{$\Delta$\%}Acc, {$\Delta$\%}E} & \textbf{Acc.} & \textbf{{$\Delta$\%}Acc, {$\Delta$\%}E} \\
\hline
\textbf{Computer Science} & 0.38 & 78,556 kJ & 0.56 & (+47\%, +42\%) & 0.39 & (+3\%, +118\%) & 0.43 & (+13\%, +13,858\%) \\
\textbf{Economics} & 0.40 & 80,437 kJ & 0.62 & (+51\%, +65\%) & 0.42 & (+5\%, +177\%) & 0.42 & (+5\%, +13,211\%) \\
\textbf{Engineering} & 0.37 & 74,805 kJ & 0.75 & (+99\%, +36\%) & 0.44 & (+19\%, +72\%) & 0.43 & (+17\%, +12,233\%) \\
\textbf{Health} & 0.50 & 78,484 kJ & 0.78 & (+57\%, +44\%) & 0.54 & (+8\%, +108\%) & 0.55 & (+10\%, +14,339\%) \\
\textbf{Humanities} & 0.44 & 79,029 kJ & 0.72 & (+61\%, +63\%) & 0.46 & (+5\%, +174\%) & 0.45 & (+2\%, +13,334\%) \\
\textbf{Math} & 0.11 & 83,532 kJ & 0.39 & (+350\%, +37\%) & 0.11 & (+0\%, +88\%) & 0.31 & (+281\%, +15,132\%) \\
\textbf{Natural Sciences} & 0.26 & 76,172 kJ & 0.54 & (+100\%, +41\%) & 0.27 & (+4\%, +102\%) & 0.29 & (+11\%, +15,483\%) \\
\textbf{Sociology} & 0.47 & 76,673 kJ & 0.82 & (+72\%, +40\%) & 0.48 & (+2\%, +97\%) & 0.60 & (+26\%, +13,198\%) \\
\hline
\end{tabular}
\caption{Accuracy and energy consumption (in kJ) of the Llama 1B model processing the MMLU benchmark. This table presents accuracy and percentage increase in accuracy and energy consumption changes when using the Llama 8B or applying Chain-of-Thought (CoT) and Majority Voting to Llama 3.2 1B, compared to the Zero-Shot Llama 1B.}
\label{tab:testtime}
\end{table*}

A simple way to improve the model quality during test-time is to aggregate multiple outputs of an SLM, called majority voting or self-consistency decoding \cite{wang2022self}. The approach is straightforward: for a given problem, we generate multiple candidate solutions and select the most frequent answer. In our experimental setting, we sampled up to 16 candidate solutions and selected the most common answer, similiar as done by Huggingface \cite{beeching2024scalingtesttimecompute}.
For Chain-of-Thought Prompting we used the same schemata as Meta published in their release of the LLaMa 3.2 family to evaluate on the MATH500 dataset~\footnote{Dataset Math500, https://huggingface.co/datasets/meta-llama/Llama-3.2-1B-Instruct-evals/viewer/Llama-3.2-1B-Instruct-evals\_\_math\_\_details}. We allowed to reason up to 5 steps and generate a maximum of up to 512 token.
These strategies are applied exclusively to the smaller variant of the LLaMA 3.2 architecture (1B parameters). The goal is to determine whether advanced test-time compute methods can enable smaller models to approach the accuracy of their larger counterparts, and in which costs.

The MMLU dataset was used as the evaluation benchmark, with each task designed to answer with only a single output token, A, B, C or D. This controlled environment ensured that the observed differences in performance and energy consumption stemmed solely from model and the test-time compute strategy.Table \ref{tab:testtime} presents a comparative analysis of the accuracy and energy consumption of different Llama models across various MMLU categories. The baseline model, Llama 1B, is compared against two variants: Majority Voting (MV) and Chain-of-Thought (CoT), along with the Llama 8B model. The percentage changes in accuracy and energy consumption relative to the Llama 1B baseline, provide insights into the efficiency and effectiveness of different inference strategies. Majority Voting (MV) slightly improves accuracy, with increases ranging from +0\% (Math) to +19\% (Engineering). The method is particularly effective in Health (+8\%), Economics (+5\%), Computer Science (+5\%), and Natural Sciences (+4\%). However, the trade-off is a significant rise in energy consumption, ranging from +72\% (Engineering) to +177\% (Economics). This suggests that while MV can offer modest accuracy improvements, it comes at a steep energy cost, making it less practical for efficiency-sensitive applications.

In contrast, Chain-of-Thought (CoT) prompting significantly improves accuracy, particularly in Math (+281\%), Sociology (+26\%), Engineering (+17\%), and Computer Science (+13\%). However, it has marginal effects in Humanities (+1\%), Economics (+5\%), and Natural Sciences (+1\%), suggesting that step-by-step reasoning is more beneficial for structured problem-solving tasks than for general knowledge-based ones. The major downside of CoT is its immense computational cost, leading to a 120x to 150x increase in energy consumption, making it highly inefficient for real-world deployment.

On the other hand, Llama 8B consistently outperforms all Llama 1B variants, with accuracy improvements ranging from 47\% (Computer Science) to 350\% (Math) while maintaining a more moderate energy increase of 35-65\%. This makes Llama 8B significantly more energy-efficient than CoT-enhanced Llama 1B, as it provides higher accuracy at a much lower relative energy cost. The model particularly excels in math-heavy and structured reasoning tasks (e.g., Math: +350\%, Sociology: +72\%), reinforcing the idea that larger models handle complex reasoning better.

This evaluation shows that for research question 3: \textit{How do test-time compute strategies, such
as Majority Voting and Chain-of-Though Prompting, affect the trade-off between accuracy and
energy consumption?} CoT Prompting, despite its accuracy benefits, is highly energy-inefficient, while Llama 8B offers a better trade-off between accuracy and energy consumption. To be highlighted is the effect on specific categories. While it makes sense to apply CoT for structured problem solving tasks, step-by-step reasoning is not beneficial for general knowledge-based tasks. This highlights the need for selective application of reasoning techniques to balance accuracy and efficiency.


\section{Discussion: A Solution for Energy-Efficient Query Routing}
\label{sec:discussion}

We highlight the fundamental trade-off between accuracy and energy efficiency for five different LLMs. 
Our analysis shows that techniques like Majority Voting provide negligible accuracy improvements with additional energy cost, whereas Chain-of-Thought (CoT) prompting significantly enhances accuracy in reasoning-heavy tasks but comes with a massive energy overhead. 
Meanwhile, Llama 8B yields substantial accuracy improvements at a more moderate energy increase, making it a more efficient alternative to CoT-enhanced Llama 1B.
Given these insights, we propose an adaptive routing mechanism to balance accuracy and energy consumption dynamically.

\begin{figure}[h]
\centering
  \includegraphics[width=\columnwidth]{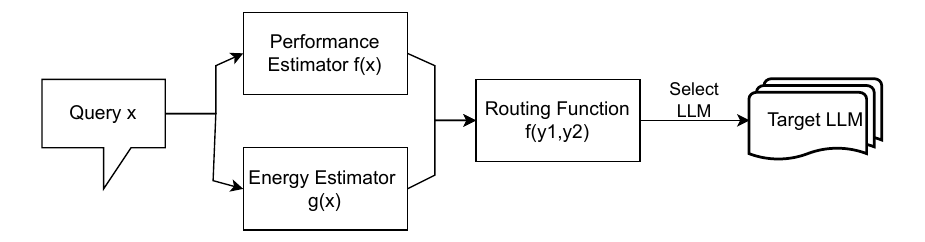}
  \caption{Routing architecture with Performance and Energy Estimator}
  \label{fig:routingArchitecture}
\end{figure}
Figure \ref{fig:routingArchitecture} visualizes an architectural design to balance energy efficiency with accuracy. 
The architecture includes two estimators:\\
\begin{itemize}
    \item \textbf{Performance Estimator}: Uses query features to predict the expected accuracy of different models and techniques.
    \item \textbf{Energy Estimator:} Estimates the energy consumption for each inference method based on historical data and model/hardware-specific energy curves.
\end{itemize}

The system dynamically selects the optimal LLM for each query, balancing accuracy and energy efficiency based on pre-collected benchmarking data. For low-complexity queries, such as those in Humanities, Natural Sciences, and Economics, where CoT provides minimal accuracy improvements, the system defaults to Llama 1B to save energy. Majority Voting (MV) can be applied in cases of uncertainty with accepting additional energy costs. For complex reasoning tasks, such as Math, Engineering, and Computer Science, the system routes queries to Llama 8B instead of using CoT on Llama 1B, as Llama 8B offers similar or better accuracy with lower energy consumption (~40-60\%). For high-accuracy critical queries, CoT remains an option, particularly in Math, where it boosts accuracy by 281\%. 

However, the system dynamically limits reasoning steps based on query complexity to prevent unnecessary energy expenditure. A key component of the approach is leveraging operating curves for token generation, which optimize energy usage in token-intensive processes like CoT. These curves provide insights into diminishing returns, highlighting when additional tokens contribute little to accuracy while significantly increasing energy costs.

By integrating token budget mechanisms, similiar as done in Sprout ~\cite{li2024sustainablegenaiusinggeneration}, the system dynamically regulates reasoning steps in CoT, ensuring only the necessary depth of reasoning is applied per query. It also predicts optimal stopping points to avoid high computational costs without sacrificing accuracy. Additionally, by adjusting token budgets based on real-time factors like server load, latency constraints, and regional carbon intensity, the system ensures sustainable inference while maintaining a balance between accuracy and energy efficiency. The challenges include accuracy-to-energy optimization, constraint enforcement, and dynamic adaptation based on conditions like GPU load or regional carbon intensity. The system’s success is measured through metrics such as energy savings and accuracy compared to using a single fixed LLM.

\section{Conclusion}
This paper evaluated the energy efficiency of different Large Language Models (LLMs) in token processing and generation, proposing an intelligent routing mechanism to balance accuracy and energy consumption. Our analysis revealed a fundamental tradeoff: while reasoning techniques such as Chain of Thought (CoT) significantly enhance the accuracy of Small Language Models (SLMs), they do so at an extreme energy cost—up to 130-150× compared to the baseline model without reasoning. These findings challenge the current trend of improving SLMs primarily through complex reasoning strategies—raising the question: at what cost? This calls for a shift toward more energy-efficient LLMs rather than relying on computationally expensive inference-time techniques.

The inference cost of LLMs is largely driven by autoregressive token generation, making energy efficiency a crucial factor in real-world applications. While high-performance LLMs are often overused for simple tasks that smaller models can handle, our findings suggest that SLMs, when selectively enhanced with reasoning techniques, can sometimes approach the accuracy of larger models for complex tasks like math-solving. However, the variability across task categories highlights the need for an adaptive routing system that intelligently selects the optimal model and reasoning approach per query.

To address this, we proposed a dynamic decision-making architecture that routes queries based on estimated complexity, accuracy requirements, and energy constraints. By leveraging operation curves—which map energy consumption to task-specific accuracy—we can systematically decide when to scale up to a larger LLM, rely on a base SLM, or use reasoning techniques like CoT. The controlled reasoning process in CoT, governed by these operating curves, enables the system to optimize token generation, ensuring that only the necessary reasoning depth is applied per query, thus minimizing excessive energy consumption. This vision shifts the focus from indiscriminate model scaling to intelligent, task-aware LLM utilization, ensuring both sustainability and efficiency in future AI applications.

\bibliographystyle{ACM-Reference-Format}
\bibliography{sample-base}

\appendix

\end{document}